\documentclass{article}
\usepackage{graphicx}
\usepackage{subfigure}
\usepackage{float}
\usepackage{booktabs}
\usepackage{url}
\usepackage{hyperref}
\usepackage{authblk}
\usepackage{amsmath}
\usepackage{textcomp,mathcomp}

\title{Root Cause Analysis on Energy Efficiency with Transfer Entropy Flow}
\author{Jian Ma}
\affil{\normalsize Hitachi China Research Laboratory\\majian@hitachi.cn}
\date{}

\newtheorem{definition}{\textbf{Definition}}
\newtheorem{theorem}{\textbf{Theorem}}

\newtheorem{corollary}{\textbf{Corollary}}

\begin{document}
\maketitle

\begin{abstract}
\noindent
Energy efficiency is a big concern in industrial sectors. Finding the root cause of anomaly state of energy efficiency can help to improve energy efficiency of industrial systems and therefore save energy cost. In this research, we propose to use transfer entropy (TE) for root cause analysis on energy efficiency of industrial systems. A method, called TE flow, is proposed in that a TE flow from physical measurements of each subsystem to the energy efficiency indicator along timeline is considered as causal strength for diagnosing root cause of anomaly states of energy efficiency of a system. The copula entropy-based nonparametric TE estimator is used in the proposed method. We conducted experiments on real data collected from a compressing air system to verify the proposed method. Experimental results show that the TE flow method successfully identified the root cause of the energy (in)efficiency of the system.\\
 
\end{abstract}

\section{Introduction}
Climate change is an urgent challenge faced by human society. To tackle it, world needs reducing carbon emission by replacing fossil fuels with renewable energies. Addition to that, we should also consider improving energy efficiency of our economy. According to the IEA's report \cite{Campbell2014}, the global economy could increase by \$18 trillion by 2035 if energy efficiency is adopted as the "first choice" for new energy supplies, which would help to achieve the goal of the Paris Agreement that limits global warming to 2 \textcelsius. UN has launched environment program for energy efficiency covering various sectors, such as building, transport, energy, manufacturing, etc. Energy efficiency is one of the main goals of the Industrial 4.0 \cite{Mourtzis2022}. As the digitization of the manufacturing systems, there is a big opportunity to improve the energy efficiency of different type of industrial equipment used in production lines. 

Anomaly detection is a common approach in machine learning that find the abnormal data among normal data \cite{Braei2020,Nassif2021}. It can be applied to improve energy efficiency by detecting the anomaly states of energy efficiency of equipment and then making intervention on them \cite{Wang2020,Himeur2021}. To achieve this goal, one should find the root cause of anomaly state in equipment systems \cite{Rooney2004,Wang2016}. Root Cause Analysis (RCA) is an approach for discovering such causal relationship from industrial process data. Due to the complicated structure of industrial systems, finding such cause is not easy. Some previous methods for root cause analysis, such as autoencoder \cite{Roelofs2021} and SEMs \cite{Qafari2020} are all based on certain model assumptions which is unrealistic in real cases. 

Transfer Entropy (TE) \cite{Schreiber2000} is an model-free information theoretical concept for measure causality. It is defined from the famous Wiener’s principle which states that if a random variable X can improve the prediction of Y then X is a cause of Y. Granger Causality (GC) \cite{Granger2001} is a measure of causality define from this principle for linear vector autoregression (VAR) models. TE can be considered as the nonlinear generalization of GC. TE is model-free and hence is an ideal tool for finding the causal relationships between variables in manufacturing systems. However, estimating TE is a notoriously difficult problem for the applications of TE. Copula Entropy (CE) is a recently proposed mathematical concept for measuring statistical independence \cite{Ma2011}. Ma proved that TE can be represented with only CE and hence proposed a non-parametric TE estimator based on such representation \cite{Ma2019}.

In this paper, we will propose to use TE for root cause diagnosis for energy efficiency. To our knowledge, this is the first time to apply TE to root cause analysis on energy efficiency. However, TE cannot be applied directly to such problem because of the non-stationary of manufacturing process. To tackle such issue, we propose a method, called TE flow, which derive a series of TE along timeline as the causality states of the energy efficiency of industrial process. In the proposed method, we will estimate the TEs in a time window with respect to a time lag horizon with the CE-based TE estimator, and then use the maximum of TEs as the causal strength in this time window for root cause diagnosis. 

As is well known, compressing air is a major source of energy in manufacturing sectors and CAS is one of the most widely used industrial system in manufacturing \cite{Benedetti2017}. CASs account for about 10\% of total industrial-energy use worldwide and meanwhile, CAS is also well known for its low energy efficiency \cite{Mousavi2014}. Improving energy efficiency of CAS is a big concern of end-users in manufacturing sectors and will save large amount of energy cost for end-users. In this work, we will apply the proposed method to a compressing air system (CAS) to diagnose the root cause of its state of energy efficiency. 

This paper is organized as follows: Section \ref{sec:related} presents the related work, the methodology on estimating TE via CE will be introduced in Section \ref{s:method}, the TE flow method will be proposed in Section \ref{sec:tef}, experiments on a CAS will be presented in Section \ref{s:exp}, finally Section \ref{sec:con} concludes the paper.

\section{Related Work}
\label{sec:related}
There are many existing methods for RCA on the related applications. For example, Roelofs et al. \cite{Roelofs2021} proposed to use Autoencoder for anomaly root cause diagnosis in wind turbines. Qafari et al. \cite{Qafari2020} proposed to use Structural Equation Models (SEMs) for RCA in industrial process. However, these methods are based on model assumptions which are usually unrealistic in real applications. Wang et al. \cite{Wang2023} proposed to use dilated convolutional neural network to do RCA on industrial process fault. However, deep learning methods have two main drawbacks: one is data intensive, and the other is lack of interpretability, which make them unsuitable for real industrial applications on RCA.

Transfer Entropy (TE) is an information theoretical concept for measuring causal relationships \cite{Schreiber2000}. It can be applied to any system without any assumptions because it is model-free. It has been applied to RCA for other problems in industrial applications \cite{Bauer2007,Yu2015,Hu2017}. CE-based TE estimator has been applied to many different problems, such as fault root cause diagnosis on hot strip mill process \cite{Dong2023}, chemical processes \cite{Pan2023,Bi2023} and satellite systems \cite{Liu2022,Zeng2022}, indoor air temperature prediction in HVAC system \cite{Li2022,Li2022a}, sentiment propagation in social systems \cite{Han2022,Zhang2022}.

We will study the energy efficiency of CAS in this paper. There are many empirical measures for improving the energy efficiency of CAS \cite{Mousavi2014}. Recently, data-driven method for energy efficiency of CAS has been gaining momentum, please refer to \cite{Thabet2020} for a review on this topic.

\section{Methodology}
\label{s:method}
\subsection{Theory of Copula Entropy}
\label{s:CopEnt}
Copula theory is about the representation of multivariate dependence with copula function \cite{joe2014,nelsen2007}. At the core of copula theory is Sklar theorem \cite{sklar1959} which states that multivariate probability density function can be represented as a product of its marginals and copula density function which represents dependence structure among random variables. Such representation separates dependence structure, i.e., copula function, with the properties of individual variables -- marginals, which make it possible to deal with dependence structure only regardless of joint distribution and marginal distribution. This section is to define an statistical independence measure with copula. For clarity, please refer to \cite{Ma2011} for notations.

With copula density, Copula Entropy is define as follows \cite{Ma2011}:
\begin{definition}[Copula Entropy]
	\label{d:ce}
	Let $\mathbf{X}$ be random variables with marginal distributions $\mathbf{u}$ and copula density $c(\mathbf{u})$. CE of $\mathbf{X}$ is defined as
	\begin{equation}
	H_c(\mathbf{X})=-\int_{\mathbf{u}}{c(\mathbf{u})\log{c(\mathbf{u})}}d\mathbf{u}.
	\end{equation}
\end{definition}

In information theory, MI and entropy are two different concepts \cite{Cover1999}. In \cite{Ma2011}, Ma and Sun proved that they are essentially same -- MI is also a kind of entropy, negative CE, which is stated as follows: 
\begin{theorem}
	\label{thm1}
	MI of random variables is equivalent to negative CE:
	\begin{equation}
	I(\mathbf{X})=-H_c(\mathbf{X}).
	\end{equation}
\end{theorem}
\noindent
The proof of Theorem \ref{thm1} is simple \cite{Ma2011}. There is also an instant corollary (Corollary \ref{c:ce}) on the relationship between information of joint probability density function, marginal density function and copula density function.
\begin{corollary}
	\label{c:ce}
	\begin{equation}
	H(\mathbf{X})=\sum_{i}{H(X_i)}+H_c(\mathbf{X}).
	\end{equation}
\end{corollary}
The above results cast insight into the relationship between entropy, MI, and copula through CE, and therefore build a bridge between information theory and copula theory. CE itself provides a mathematical theory of statistical independence measure.

\subsection{Estimating Copula Entropy}
\label{s:est}
It has been widely considered that estimating MI is notoriously difficult. Under the blessing of Theorem \ref{thm1}, Ma and Sun \cite{Ma2011} proposed a simple and elegant non-parametric method for estimating CE (MI) from data which comprises of only two steps\footnote{The \textsf{R} package \textsf{copent} \cite{Ma2020} for estimating CE is available on CRAN and also on GitHub at \url{https://github.com/majianthu/copent}.}:
\begin{enumerate}
	\item Estimating Empirical Copula Density (ECD);
	\item Estimating CE.
\end{enumerate}

For Step 1, if given data samples $\{\mathbf{x}_1,\ldots,\mathbf{x}_T\}$ i.i.d. generated from random variables $\mathbf{X}=\{x_1,\ldots,x_N\}^T$, one can easily estimate ECD as follows:
\begin{equation}
F_i(x_i)=\frac{1}{T}\sum_{t=1}^{T}{\chi(\mathbf{x}_{t}^{i}\leq x_i)},
\end{equation}
where $i=1,\ldots,N$ and $\chi$ represents for indicator function. Let $\mathbf{u}=[F_1,\ldots,F_N]$, and then one can derive a new samples set $\{\mathbf{u}_1,\ldots,\mathbf{u}_T\}$ as data from ECD $c(\mathbf{u})$. In practice, Step 1 can be easily implemented non-parametrically with rank statistic.

Once ECD is estimated, Step 2 is essentially a problem of entropy estimation which has been contributed with many existing methods. Among them, the kNN method \cite{Kraskov2004} was suggested in \cite{Ma2011}. With rank statistic and the kNN method, one can derive a non-parametric method of estimating CE, which can be applied to any situation without any assumption on the underlying system.

\subsection{Estimating Transfer Entropy via Copula Entropy}
\label{sec:tevce}
Transfer Entropy (TE) \cite{Schreiber2000} is another information theoretical based concept for measuring causal relationship between time series. 

Given two time series $X_i$ and $Y_i, i=1\ldots,T$, the TE from $X$ to $Y$ is defined as
\begin{equation}
	TE_{x\rightarrow y}=\sum_{i}{p(Y_{i+1},Y^i,X_i)\log{\frac{p(Y_{i+1}|Y^i,X_i)}{p(Y_{i+1}|Y^i)}}},
\end{equation}
where $p()$ is for joint probabilistic density and $Y^i=(Y_1,\cdots,Y_i)$.

Ma \cite{Ma2019} proved that TE can be represented with only CE, as follows:
\begin{equation}
	TE_{x\rightarrow y}=-H_c(Y_{i+1},Y^i,X_i)+H_c(Y_{i+1},Y^i)+H_c(Y_{i+1},X_i)-H_c(Y^i).
	\label{eq:tece}
\end{equation}
Based on this representation, a nonparametric estimator of TE was proposed based on the nonparametric estimator of CE \cite{Ma2019}, which composed of two simple steps: first estimating the CE terms in \eqref{eq:tece} with the nonparametric CE estimator and then calculating TE from the estimated CE terms.

\section{Transfer Entropy Flow}
\label{sec:tef}
In this research, we study a type of system which is assumed to be composed of a group of parallel subsystems. The output of the whole system is a combination of that of all the subsystems. During the running of the system, we monitor the state of energy efficiency of the system. We then want to find the root cause of energy inefficiency of the system, i.e., which subsystem causes such inefficiency. TE can be utilized for discovering such causal relationships. However, due to non-stationary of systems, TE cannot be applied to target system directly. 

To tackle this issue, we propose a TE-based method for diagnosing energy inefficiency of such system, called Transfer Entropy Flow (TE flow), which composed of the following steps:

\begin{enumerate}
	
\item divide the time series data collected from the system by a group of consecutive time windows;

\item compute the indicator of energy efficiency of the whole system as the ratio of the output energy to input energy of the system in each time window;

\item estimate the TEs with different time lags from the energy used by each subsystem to the indicator of energy efficiency of the whole system;

\item find the maximum of the estimated TEs as the causal strength from subsystem to energy efficiency in each time window.

\end{enumerate}

The output of the TE flow method are time series of the maximum of TE values corresponding to the causal strength in every time windows for the subsystems. These results can be used for diagnosing the root cause of energy efficiency of the system as time changes.

In the proposed method, the non-parametric TE estimator in Section \ref{sec:tevce} is adopted for estimating TEs in each time window in the TE flow method. Since TE is model-free and the TE estimator used is nonparametric, the proposed method makes no assumption on the underlying dynamical system and therefore universally applicable.

\section{Experiments and Results}
\label{s:exp}

\subsection{The system}
The target system in this study composes of 6 compressors and 6 dryers controlled with a group of control policies so as to supply compressing air with stable pressure. The 6 compressors here include 2 variable speed compressor and 4 fixed speed compressors. The output of each compressor is connected to a dryer with pipes and after processed by driers, the compressed airs are then transported through pipes to an air storage pot as the output of the whole system for energy use. 

\subsection{Data}
We aim to improve the energy efficiency of this CAS system. For this purpose, we collected two working data of the system during May 25th $\sim$ 28th and September 21th $\sim$ 22th, 2022 respectively. The data collected include the current, inlet and outlet temperature and pressure of each compressor, and the outlet flow rate and pressure of the air storage pot. The sample rate is 0.1 sample per second. During data collection, only 3 compressors, including one variable speed compressor (compressor 1) and two fixed speed compressor (compressor 3 and 6), were running for energy supply.

\subsection{Experiments}
We applied the TE flow method to the above collected data to study the energy efficiency of the compressing air system studied. For this aim, we applied our method to find the causal relationships from each compressor to the energy efficiency of the whole system.

For the first step of our method, the collected data are separated into time windows and the length of each time windows is 30 minutes which contains 180 samples. 

For the second step of our method, we use the current of each compressor as its energy cost and the total current of all the compressor as the whole energy cost of the CAS. The outlet flow rate is usually taken to measure the output energy of the whole system. We calculate the product of the flow rate and pressure as the measure of the output energy of the whole system since the pressure is not a constant due to the fluctuated need of compressing air and therefore flow rate cannot be taken as the measure of output energy alone as usual. Then we derived the energy efficiency indicator of the whole system by dividing the output energy, i.e., the product of flow rate and pressure of the output of the system, by the input energy, i.e., the total current of the whole system. With this indicator, the normal and anomaly state of energy efficiency of the system can be measured and differentiated. 

For the third step of our method, we estimate the TEs with different time lags from the current of each compressor to the indicator of energy efficiency of the whole CAS in each of a group of non-overlapping consecutive time windows. In each time window, we estimate the TE of 36 consecutive time lags (6 minutes).

In the fourth step of our method, the maximum of TEs of different time lags in each time window will be considered as the causal strength from each subsystem to energy efficiency. The estimated maximum TEs are considered as the causal contribution of each subsystem to the energy efficiency of the whole system. If there is energy inefficiency, the subsystem with high TE values will be identified as root cause accordingly. After this step, we derived the TE flows of each subsystem for diagnosing the state of energy efficiency of the system along the timeline. 

In the experiments, the above TE flow method were applied to the two group of data collected from the CAS during two working periods. 

\subsection{Results}
\begin{figure}
	\centering
	\subfigure[TE flow]{
		\includegraphics[width=\linewidth]{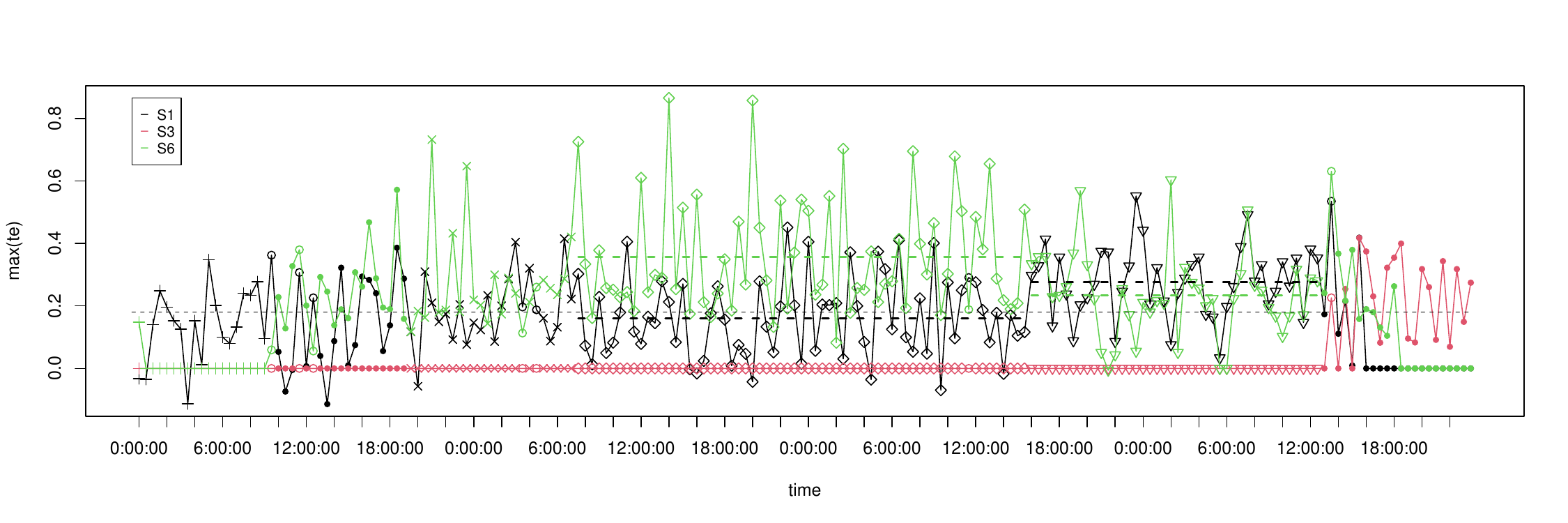}
	}
	\subfigure[Working currents]{
		\includegraphics[width=\linewidth]{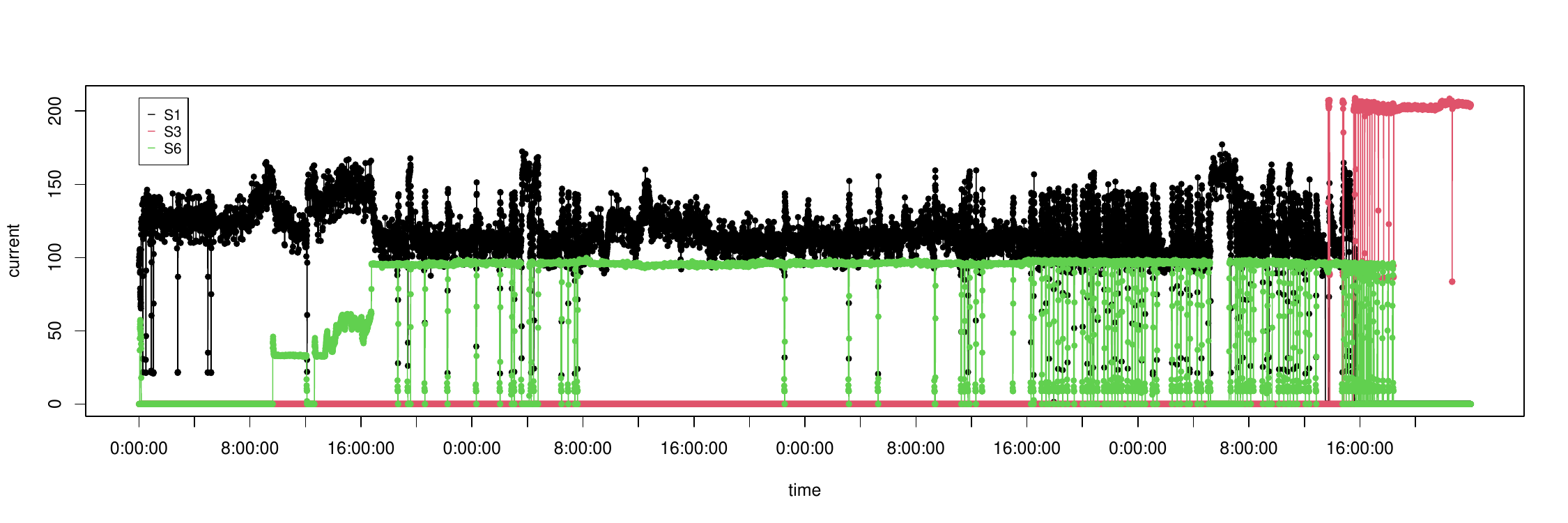}
	}
	\subfigure[energy efficiency indicator]{
		\includegraphics[width=\linewidth]{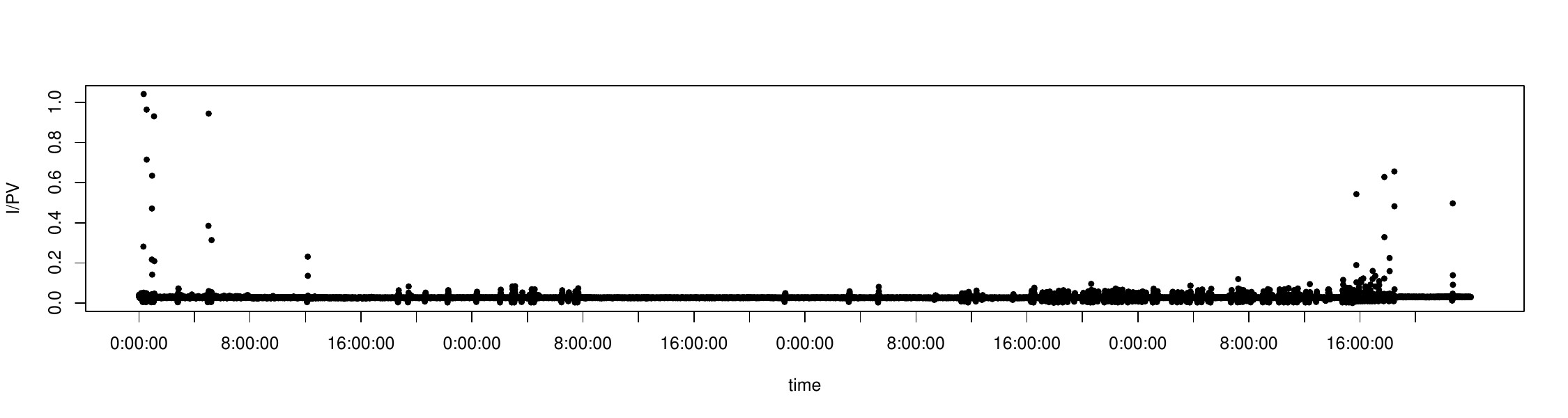}
	}
	\caption{Results of Case 1.}
	\label{fig:case1}
\end{figure}

\begin{figure}
	\centering
	\subfigure[TE flow]{
		\includegraphics[width=\linewidth]{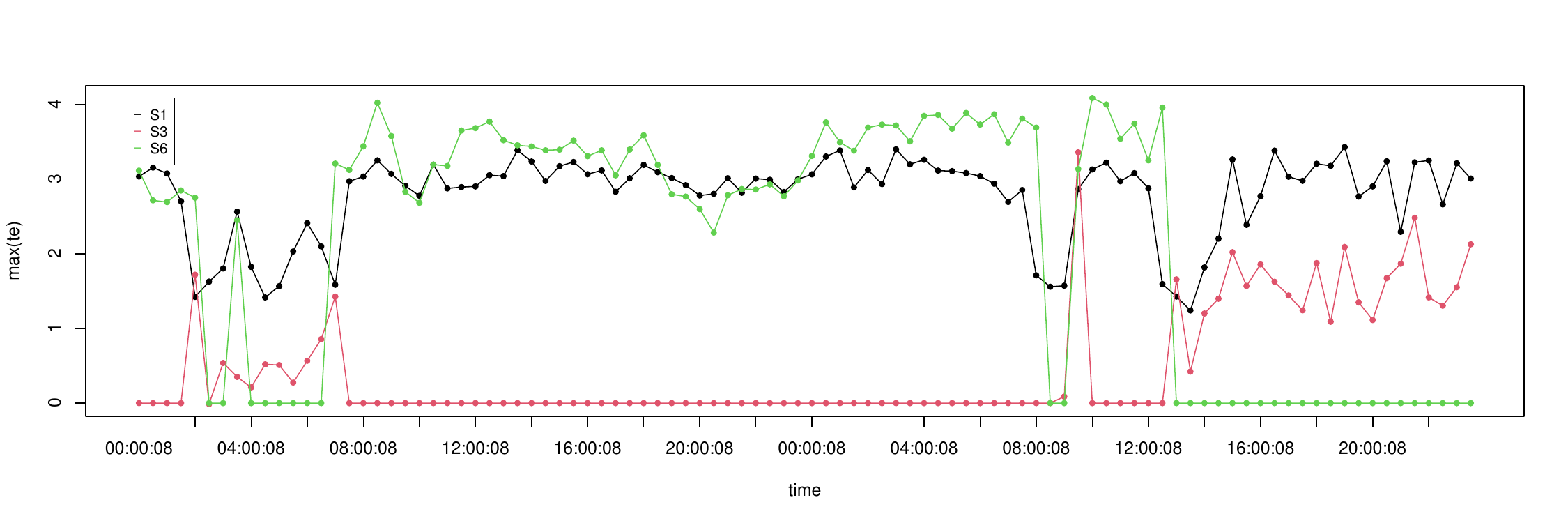}
	}
	\subfigure[Working currents]{
		\includegraphics[width=\linewidth]{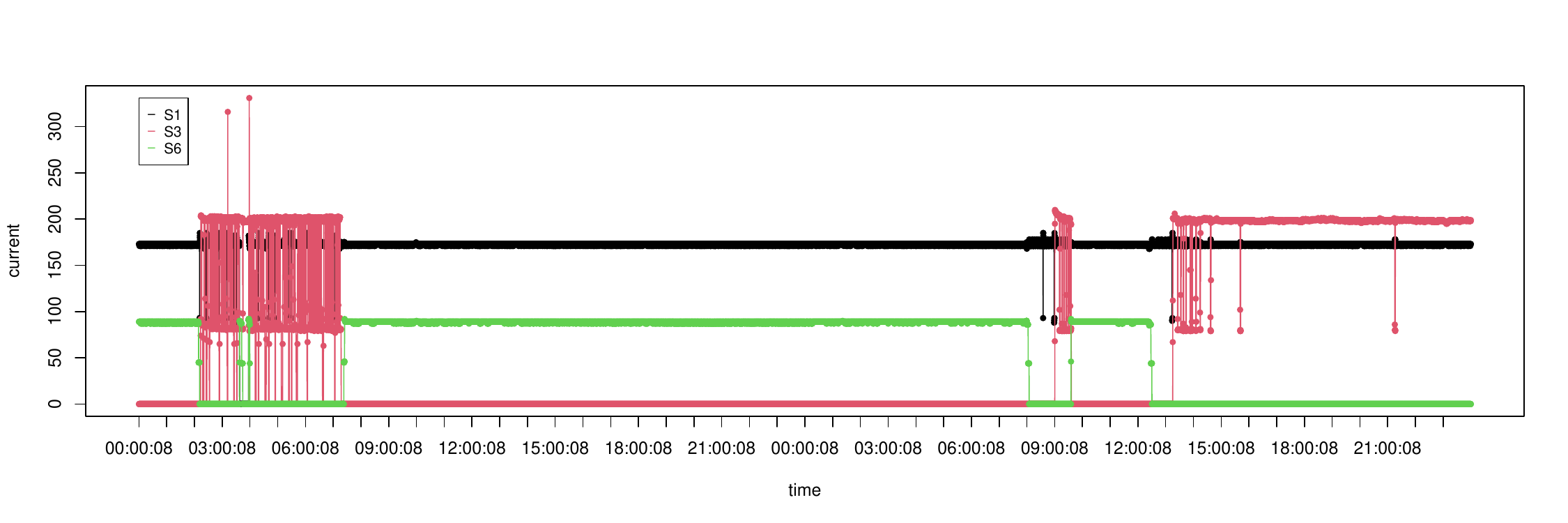}
	}
	\subfigure[energy efficiency indicator]{
		\includegraphics[width=\linewidth]{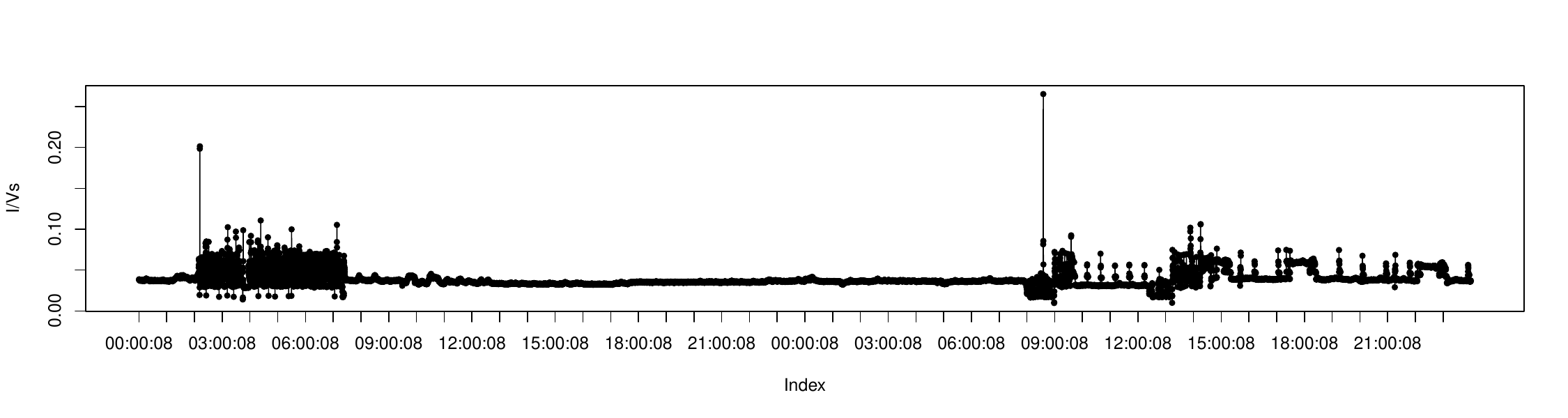}
	}
	\caption{Results of Case 2.}
	\label{fig:case2}
\end{figure}

The results of the real data experiments are shown in Figure \ref{fig:case1} and \ref{fig:case2}, each of which includes the result of the TE flow method along with the current of the subsystems and the indicator of energy efficiency of the whole system. In the results of the TE flow methods, the estimated TE flows of the three running subsystems are shown in one plot jointly, from which one can learn that 1) the values of TE flow reflect the running state (currents) of the compressors; 2) the trends of TE flow change as the currents of the compressors. 

\paragraph{Case 1}
In the results of the data during May 25th $\sim$ 28th, 2022 (see Figure \ref{fig:case1}), the system is running efficiently during the middle time window (8h,26th $\sim$ 16h,27th) as shown in the energy efficiency indicator and in the current. Meanwhile, we can learn from the result of the estimated TE flow that the TE flow of compressor 6 is relatively larger than that of compressor 1. This can be interpreted that the energy efficiency is mainly caused by the running of compressor 6 rather than compressor 1. In another time window (16h,27th $\sim$ 12h,28th) of the same data when the system is running inefficiently and in abnormal state as identified with the indicator and also judged by experts, the TE flow of compressor 1 is relatively larger than that of compressor 6. This results are reasonable because the system is energy efficient if fixed speed compressor (compressor 6) has priority in control and therefore larger TE, than variable speed compressor (compressor 1) for energy supply, vice versa.

\paragraph{Case 2} In the results of another data during Sept. 21th $\sim$ 22nd, 2022 (see Figure \ref{fig:case2}), it can be learned that during 8h,21th $\sim$ 6h,22nd, the system runs at full load since the current of compressor 1 and compressor 6 reach their maximum constantly and therefore the system is running efficiently as shown by the indicator. One can learn from the corresponding results of the TE flow method that the TE value of compressor 1 and compressor 6 are with a similar constant trend, which can be interpreted as that both systems are the root causes of such full-load efficiency.

\section{Conclusions}
\label{sec:con}
In this research, we propose to apply TE to root cause analysis on industrial energy efficiency. A method, called TE flow, is proposed for such analysis which composed of four steps. The subsystem with high TE values estimated with the proposed method will be considered as the root cause of energy (in)efficiency of a system. Since the TE estimator used in the method is based on the nonparametric CE estimator, the proposed method can be applied to any systems without assumptions. We applied the proposed TE flow method to a CAS. Experimental results show that the proposed method can identify the root cause of the working states of energy (in)efficiency of the system. 

In the future, we will apply the TE flow method to more real cases. It is also interesting if some rules can be summarized from the results of the TE flow method for explaining and improving energy efficiency. This will be a future research direction.

\bibliographystyle{unsrt}
\bibliography{rca}

\end{document}